\documentclass[letterpaper]{article}
\usepackage{aaai2026}
\usepackage{times}
\usepackage{helvet}
\usepackage{courier}
\usepackage[hyphens]{url}
\usepackage{graphicx}
\usepackage{natbib}
\usepackage{caption}
\usepackage{amsmath}
\usepackage{newfloat}
\usepackage{enumitem}

\title{Unified Interactive Multimodal Moment Retrieval via Cascaded Embedding-Reranking and Temporal-Aware Score Fusion}

\author {
    Thanh Toan Le Ngo\textsuperscript{\rm 1,\rm 5},
    Huu Phat Ha\textsuperscript{\rm 1,\rm 5},
    Duy Tan Nguyen Dang\textsuperscript{\rm 4},
    Minh Thong Nguyen Le\textsuperscript{\rm 2,\rm 5},
    Tinh Anh Nguyen Nhu\textsuperscript{\rm 3,\rm 5},
}
\affiliations {
    \textsuperscript{\rm 1}University of Information Technology, Ho Chi Minh City, Vietnam\\
    \textsuperscript{\rm 2}International University, Ho Chi Minh City, Vietnam\\
    \textsuperscript{\rm 3}Ho Chi Minh City University of Technology, Ho Chi Minh City, VietNam\\
    \textsuperscript{\rm 4}AI VIET NAM\\
    \textsuperscript{\rm 5}Vietnam National University, Ho Chi Minh City, Vietnam\\
    23521603@gm.uit.edu.vn, 22521067@gm.uit.edu.vn, nddtan2011@gmail.com, thongnlm29@mp.hcmiu.edu.vn, anh.nguyennhu2306@hcmut.edu.vn
}

\begin{document}

\maketitle

\begin{abstract}
The exponential growth of video content has created an urgent need for efficient multimodal moment retrieval systems. However, existing approaches face three critical challenges: (1) fixed-weight fusion strategies fail under cross-modal noise and ambiguous queries, (2) temporal modeling struggles to capture coherent event sequences while penalizing unrealistic gaps, and (3) systems require manual modality selection, reducing usability. We propose a unified multimodal moment retrieval system with three key innovations. First, a cascaded dual-embedding pipeline combines BEiT-3 and SigLIP for broad retrieval, refined by BLIP-2 based reranking to balance recall and precision. Second, a temporal-aware scoring mechanism applies exponential decay penalties to large temporal gaps via beam search, constructing coherent event sequences rather than isolated frames. Third, Agent-guided query decomposition (GPT-4o) automatically interprets ambiguous queries, decomposes them into modality-specific sub-queries (visual/OCR/ASR), and performs adaptive score fusion eliminating manual modality selection. Qualitative analysis demonstrates that our system effectively handles ambiguous queries, retrieves temporally coherent sequences, and dynamically adapts fusion strategies, advancing interactive moment search capabilities.
\end{abstract}

\section{Introduction}

The exponential growth of video content across multiple domains has made efficient video retrieval a critical challenge. In 2022 alone, over 500 hours of new video were uploaded every minute to online platforms \cite{navarrete2025closer}, a trend further accelerated by the emergence of new platforms such as TikTok and similar short-form video services. This content spans diverse domains---from educational lectures and tutorials to news broadcasts and entertainment creating an increasingly heterogeneous and complex video ecosystem. Moreover, each video encodes information across multiple modalities: visual scenes depicting objects and actions, spoken dialogue and background audio, and textual information appearing on-screen (e.g., captions, signs, and UI elements) \cite{wan2025clamr,nguyen2024improvinggeneralizationvisualreasoning}. Real user queries are often free-form and unclear \cite{zamani2019analyzing}. People rarely say which channel to search (visual, OCR, or ASR), and the quality of each channel can vary greatly (e.g., noisy audio, OCR mistakes). Simple, fixed fusion breaks under this ambiguity and cross-modal noise, and asking users to build queries themselves makes the system harder to use \cite{zamani2020analyzing}. This multimodal richness raises a fundamental question: How can we design a multimodal moment retrieval system that can understand and decompose user' ambiguous queries in natural language, then flexibly select and fuse modalities (visual/OCR/ASR) to return relevant results?

However, leveraging multiple modalities effectively is far from straightforward. Francis et al. \cite{francis2019fusion} demonstrated that background noise in audio tracks or erroneous OCR extractions---simple fusion strategies (e.g., averaging or Reciprocal Rank Fusion) can actually degrade retrieval performance. Alternative methods segment videos into shots or keyframes and then individually embed each unit. This fine-grained indexing-creating separate vectors for each scene or keyframe---improves retrieval of specific moments but requires processing substantially larger data volumes \cite{rossetto2021retrieval,nguyen2025hybridunifiediterativenovel}.

Sun et al. \cite{sun2020multi} emphasised the importance of jointly encoding multiple modalities. Similarly, Chen et al. \cite{chen2024verified} introduced the VERIFIED benchmark and observed that many user queries remain rather coarse-grained, indicating the need for models capable of capturing more fine-grained video semantics.

Currently, temporal modeling methods in moment retrieval can generally be split into three main categories. The first category is \textbf{Fixed Temporal Windows}, which are widely used in many Video Browser Showdown (VBS) systems , but often struggle to handle events of varying durations. The second is \textbf{Attention-based Methods}, which apply temporal attention mechanisms to assign weights across time, yet often lack explicit strategies to penalize large temporal gaps \cite{ma2022xclip}. Finally, \textbf{Simple Temporal Filters}, such as window-based matching approaches used in CLIP \cite{radford2021learning}, remain effective for coarse temporal reasoning but are limited in capturing more complex dependencies across shots \cite{nguyennhu2025stervlmspatiotemporalenhancedreference}. 

While each method offers distinct advantages, none fully solves the joint challenge of retrieving events that differ greatly in duration and rely on subtle temporal links across segments. This limitation shows the need for more flexible and context-aware temporal modeling approaches in multimodal moment retrieval.

In this paper, we propose a unified multimodal video browsing and retrieval system with an integrated processing pipeline:

\begin{itemize}[leftmargin=1.5em, labelsep=0.5em, noitemsep, topsep=2pt]
  \item \textbf{Cascaded dual-embedding retrieval pipeline:} We employ a multi-stage retrieval framework that combines BEiT-3 \cite{wang2023image} and SigLIP \cite{zhai2023siglip} for broad candidate retrieval and refines results using BLIP-2-based image-text matching, achieving a trade-off between coverage and precision.

  \item \textbf{Temporal event reasoning with exponential decay:} We propose a temporal-aware scoring mechanism that applies exponential decay to penalize large temporal gaps, allowing the system to detect coherent event sequences rather than isolated frames.

  \item \textbf{Agent-guided multimodal query decomposition and fusion:} We leverage GPT-4o to interpret ambiguous natural-language queries, decompose them into modality-specific sub-queries (visual, OCR, ASR), route them to corresponding retrieval modules, and perform adaptive score fusion without requiring users to specify retrieval modalities manually.
\end{itemize}
\begin{figure*}[t]
\centering
\includegraphics[width=1.2\columnwidth]{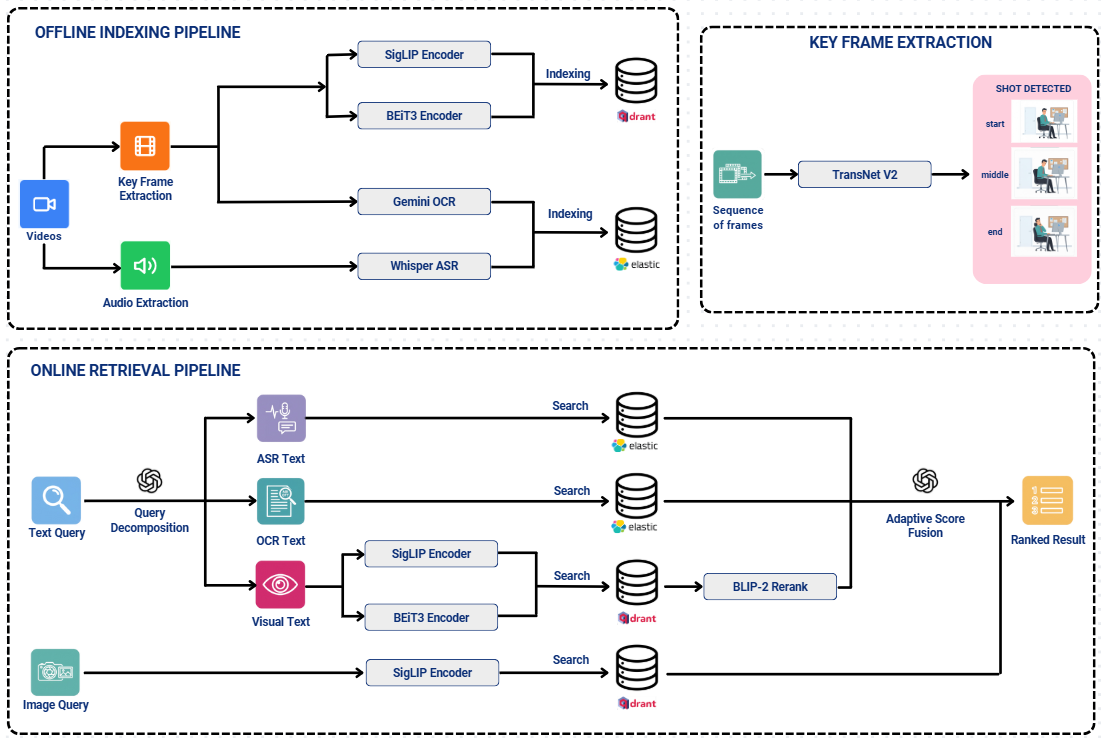}
\caption{Overview of the proposed multimodal video browsing and retrieval system.}
\label{fig:pipeline}
\end{figure*}

\section{Related Work}

Recent video retrieval systems like VISIONE~\cite{amato2024visione} get strong performance by using multimodal models (OpenCLIP~\cite{schuhmann2022laion}, CLIP2Video~\cite{fang2021clip2video}). However, most are single-stage architecture where one model handles both indexing and search. Chen et al.~\cite{chen2020fine} said that ``global embeddings struggle to capture fine-grained semantics,'' as optimizing for both recall and precision remains difficult. While BLIP-2~\cite{li2023blip2} enables effective reranking, its $\mathcal{O}(n)$ complexity~\cite{ye2024improving} limits scale.

Temporal modeling remains limited: \textbf{fixed windows}~\cite{amato2024visione} struggle with varying event durations, \textbf{attention methods}~\cite{sun2020multi} lack explicit gap penalties, and \textbf{simple filters}~\cite{francis2019fusion} cannot capture complex dependencies~\cite{ye2024improving,wang2023image,ma2022xclip,tran2025efficientrobustmomentretrieval}.

For query processing, systems like VISIONE~\cite{amato2024visione} and Dionysus~\cite{nguyen2024dionysus} require manual modality selection. Rosa \textit{et al.}~\cite{rosa2025smart} proposed smart routing via GPT-4.1, but focused on routing rather than decomposition or fusion.

We address these challenges through cascaded retrieval, temporal reasoning with exponential decay, and LLM-guided query processing.

\section{System Architecture}

Our system is composed of two primary pipelines. The first, an offline pipeline, is responsible for video processing to extract and index multimodal data (visuals, audio, OCR text). The second, the online retrieval pipeline, processes user queries to deliver ranked results ( see Figure 1 ).

\subsection{Offline Indexing Pipeline}

The offline indexing pipeline pre-processes video content to construct a searchable multimodal index. It first extracts the audio track and applies TransNetV2 \cite{soucek2020transnet} for shot segmentation and keyframe selection, and the data then flows into three parallel streams: (1) visual embeddings (BEiT-3 + SigLIP in Qdrant), (2) OCR text extraction (Gemini 2.0 Flash), and (3) ASR transcription (Whisper Large-v3).

\subsubsection{Video Pre-processing and Keyframe Extraction}

First, we separate the audio track from the video, while the visual stream is processed using the TransNetV2 \cite{soucek2020transnet} model to detect shot boundaries. For each shot, three representative keyframes are extracted.

\subsubsection{Visual Embedding Generation}

For each keyframe, we extract two dual visual embeddings using BEiT-3 \cite{wang2023image} and SigLIP \cite{zhai2023siglip}. By combining both representations, we use BEiT-3's high semantic precision and SigLIP's broad generalization capability, leading to more robust overall retrieval performance. Both embeddings are normalized and stored jointly in Qdrant using named vectors to support unified multimodal querying.

\subsubsection{OCR Text}

Extracting text from video frames is challenging because the on-screen text may appear in various styles, orientations, or may be blurred or partially occluded. Recent work has shown that multimodal large language models (MLLMs) \cite{yin2024survey} can handle such complex scene text more effectively than traditional OCR systems such as Tesseract or PaddleOCR \cite{kosugi2023ocr,ye2024improving}. Therefore, we use Google Gemini 2.0 Flash \cite{pichai2024gemini} to extract text from each keyframe using a simple JSON-based prompt. This approach allows Gemini to understand text within its visual context, handle multilingual content (e.g., English and Vietnamese), and produce clean, structured outputs suitable for indexing.

\subsubsection{ASR-Based Speech Transcription}

Audio often conveys essential information through dialogue, narration, and spoken descriptions, making it an important modality for video retrieval. We convert speech into text using Whisper Large-v3 \cite{openai2023whisper}, which is well suited for this task thanks to its strong multilingual capability. The output is divided into speech segments with precise timestamps, and each segment is treated as a semantic unit for retrieval. By aligning these timestamped segments with the nearest keyframes along the video timeline, which can then be mapped directly to the corresponding visual segment.

\subsection{Online Indexing Pipeline}

\subsubsection{Query Decomposition using agent}

In real-world interactive search scenarios, users often write queries based on \textit{vaguely remembered details}. Therefore, modern systems must infer the relevant modality and combine multiple channels to achieve high retrieval performance \cite{francis2019fusion}. Our approach, instead of requiring users to specify the modality, we employ GPT-4o to split the query into modality-specific components (visual/OCR/ASR) with corresponding weights. This method not only identifies the relevant modalities but also estimates their relative importance ( see Figure 2 ).

\begin{figure}[t]
\centering
\includegraphics[width=0.7\columnwidth]{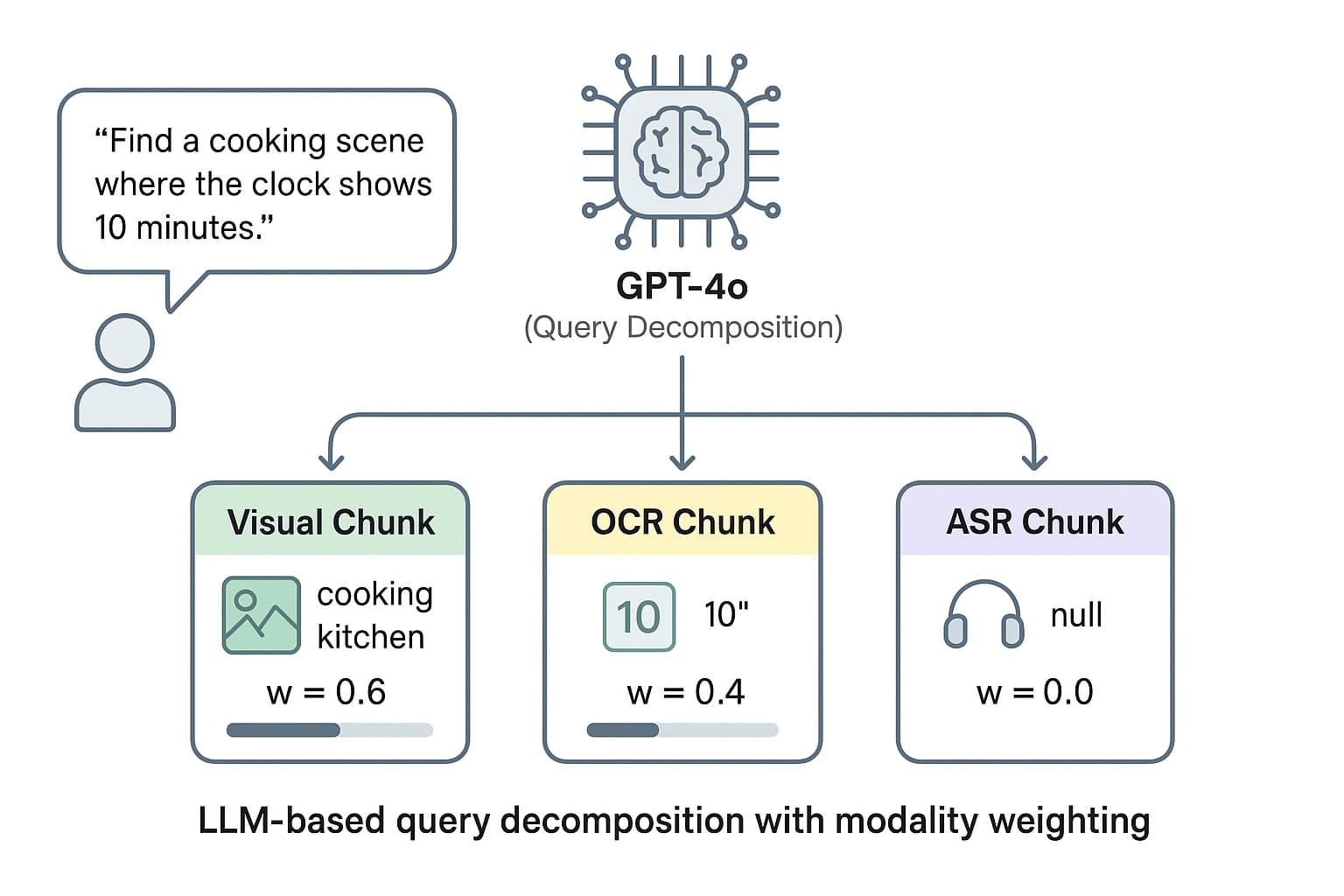}
\caption{Decomposition of a multimodal query.}
\label{fig:query_decomposition}
\end{figure}

\subsubsection{Parallel Search Strategy}
\label{sec:parallel_search}

After the query is split into modality components, the system runs independent retrieval across three branches, each employing a distinct scoring mechanism. Each branch outputs a ranked list of keyframes (or ASR segments mapped to their nearest keyframes) along with their modality-specific relevance scores. In the visual search (Qdrant) branch, each keyframe is represented by two embeddings (SigLIP \cite{zhai2023siglip} and BEiT-3 \cite{wang2023image}). Both embeddings are cosine-matched against the stored keyframes, and the results from the two models are combined using Score-Reflected Reciprocal Rank Fusion (SRRF), which preserves the original similarity scores rather than relying solely on rank positions as in standard RRF. The top-100 candidates obtained from SRRF are subsequently re-ranked using BLIP-2's \cite{li2023blip2} Image-Text Matching (ITM) head, which applies cross-modal attention to compute fine-grained semantic alignment scores between the visual query and each candidate frame. In parallel, the OCR and ASR search (Elasticsearch) branch indexes text extracted from on-screen captions (OCR) and speech transcripts (ASR) as documents. During retrieval, Elasticsearch computes relevance scores using multiple matching strategies (exact phrase, full-term match, partial match, and fuzzy match), reflecting how well the textual content aligns with the user query.

\subsubsection{Adaptive Score Fusion}

When merging results from different modalities (visual, OCR, ASR), we face two main challenges: (1) each modality produces scores on very different scales (e.g., visual similarity scores are not directly comparable to textual relevance scores from Elasticsearch), and (2) the importance of each modality depends on the query itself, making fixed weighting ineffective across all cases.

We begin by applying min–max normalization, which rescales scores to a common range while preserving their intra-modality ranking. For modality $m$ and keyframe $f$, the normalized score is given by:

\begin{equation}
s_m^{\mathrm{norm}}(f) = 
\frac{s_m(f) - \min(s_m)}{\max(s_m) - \min(s_m)+\epsilon}
\label{eq:norm}
\end{equation}

After normalization, we fuse modality scores using the agent-predicted weights. The final fusion score $S(f)$ is computed as:

\begin{equation}
S(f) = \sum_{m \in \{\mathrm{vis},\, \mathrm{ocr},\, \mathrm{asr}\}} w_m \cdot s_m^{\mathrm{norm}}(f)
\label{eq:fusion}
\end{equation}

where $w_m$ denotes the weight assigned to modality $m$, as predicted by the agent based on the query semantics. These weights reflect the relative importance of each modality for a specific search request.

This adaptive fusion strategy allows the system to dynamically adjust to different query types, leverage the strengths of each modality, and remain robust even when one modality is noisy or unavailable.

\section{Methodology}

\subsection{Cascaded Dual-Embedding Retrieval Pipeline}

Large-scale video retrieval must balance speed and ranking quality. Cross-encoders (e.g., BLIP-2 \cite{li2023blip2}) score a query–frame jointly, capturing rich cross-modal interactions; however, this is impractical at collection scale as the computational cost grows linearly with the number of pairs. Conversely, dual encoders (e.g., SigLIP \cite{zhai2023siglip}, BEiT-3 \cite{wang2023image}) support efficient retrieval via precomputed indexes, but lack the token-level cross-attention needed for fine-grained alignment.

To address this trade-off, we adopted the cascaded ``retrieval-then-rerank'' pipeline, as detailed in Section \ref{sec:parallel_search}. This layered design leverages the scalability of dual encoders (SigLIP, BEiT-3) for an efficient first-pass retrieval (optimizing recall) and reserves the precise, but costly, cross-encoder (BLIP-2 ITM) for a second-pass reranking on a small candidate set (optimizing precision). This yields a practical and effective balance between efficiency and accuracy.

\subsection{Temporal Search with Adaptive Decay and Multi-Stage Refinement}

Standard frame-level retrieval approaches struggle with two critical issues: (1) combinatorial explosion when aligning multi-event queries to candidate frames across long videos, and (2) temporal misalignment, where semantically relevant frames may be scattered across unrealistic time spans, resulting in disjointed or implausible sequences. To address these challenges, we propose Temporal Search with Adaptive Decay and Multi-Stage Refinement, an effective method that builds coherent event sequences while applying soft temporal constraints through exponential decay weighting and fine-grained post-validation.

\subsubsection{Temporal Sequence Construction via Beam Search}

To construct a coherent temporal sequence, the system must address the combinatorial complexity of aligning multiple event candidates. Given a query containing $K$ events and $M$ candidate frames per event. To mitigate this, we employ a beam search algorithm \cite{meister2020best,cohen2019empirical,lemons2022beam} that retains only the top-$B$ partial sequences (beams) at each iteration, reducing complexity from exponential to $O(B \times K \times M)$ while maintaining near-optimal solutions. This greedy approximation ensures computational tractability while preserving diversity in the search space, preventing premature convergence to suboptimal local maxima.

\subsubsection{Temporal Decay Weighting for Coherence Enforcement}

To enforce temporal realism and maintain natural event flow, we introduce an \textbf{exponential decay weighting factor} for each event transition:

\begin{equation}
\lambda_i = e^{-\alpha \cdot \Delta t_i}, \quad \Delta t_i = t_i - t_{i-1}
\end{equation}

where $\alpha$ is a hyperparameter controlling temporal sensitivity, and $\Delta t_i$ represents the time gap between consecutive events.

\textit{Rationale:} The exponential form naturally models temporal decay processes, applying a soft penalty to large temporal gaps ($\lambda_i \to 0$ as $\Delta t_i \to \infty$) while remaining tolerant of small, realistic delays ($\lambda_i \to 1$ as $\Delta t_i \to 0$). This formulation provides several advantages over alternative approaches \cite{nguyennhu2025lightweight}.

\textit{Compared to hard thresholds}, exponential decay avoids binary cutoffs that would abruptly invalidate sequences exceeding a fixed time limit, instead providing smooth degradation. \textit{Compared to linear decay}, the exponential function more accurately reflects human perception of temporal coherence, where nearby events feel strongly connected while distant events feel increasingly unrelated. \textit{Compared to ABTS's local stability measure}, while ABTS \cite{nguyennhu2025lightweight} computes temporal stability via variance within fixed neighborhoods, our global decay mechanism enforces temporal constraints across the entire sequence, complementing local frame consistency with sequence-level temporal realism. This temporal decay acts as a \textit{soft prior on event relatedness}: events occurring in quick succession (e.g., $\Delta t < 2$s) receive weights near 1.0, maintaining full scoring contribution, while large gaps (e.g., $\Delta t > 10$s) are exponentially penalized, naturally discouraging implausible temporal configurations.

\subsubsection{Sequence Scoring with Additive Aggregation}

The cumulative score $SS_j$ for a candidate sequence $j$ is computed as a weighted sum of event-wise similarity scores modulated by temporal decay:

\begin{equation}
SS_j = \sum_{i=1}^{K} s_i \cdot e^{-\alpha(t_i - t_{i-1})}
\end{equation}

The beam search algorithm selects the optimal sequence $S^*$ that maximizes this cumulative score \cite{meister2020best}:

\begin{equation}
S^* = \arg\max_{SS_j} \sum_{i=1}^{K} s_i \cdot e^{-\alpha(t_i - t_{i-1})}
\end{equation}

\textit{Rationale:} We adopt an additive scoring approach, which is more robust than multiplicative aggregation. A multiplicative formulation ($\prod_i s_i \cdot \lambda_i$) would be overly sensitive to a single low-scoring transition. The additive model allows minor weak links without invalidating the entire chain, recognizing that real-world video retrieval often involves imperfect matches where most events align well but occasional transitions may be ambiguous.

This design contrasts with ABTS's single-moment scoring \cite{nguyennhu2025lightweight}. We generalize to sequence-level scoring: a temporal decay ($\lambda_i$) multiplies each event's contribution rather than being added as a separate term, naturally down-weighting distant events while preserving overall sequence viability.

\subsubsection{Final Reranking with BLIP2-Based Validation}

The best candidate sequence $S^*$ identified by beam search undergoes a BLIP2-based post-reranking stage for fine-grained re-evaluation. The final enhanced score for each event is then computed as:

\begin{equation}
S_i^{(\text{final})} = s_i \cdot \lambda_i \cdot b_i
\end{equation}

\textit{Rationale:} This product integration acts as a gating mechanism, enforcing that high final scores occur only when all three criteria are together satisfied: \textit{Semantic relevance} ($s_i$) ensures the frame matches the query semantically. Temporal coherence ($\lambda_i$) applies exponential decay to maintain temporally plausible positioning. Fine-grained alignment ($b_i$) leverages BLIP-2 validation to confirm detailed image-text correspondence beyond coarse embeddings. If any single component is low, the overall score is suppressed, ensuring a \textit{strict multi-faceted quality constraint}. In the final validation stage, multiplicative gating enforces stringent quality requirements, significantly \textit{reducing false positives} where semantically plausible but visually misaligned frames might otherwise pass through. The final sequence score is computed as:

\begin{equation}
SS^{(\text{final})} = \sum_{i=1}^{K} s_i \cdot \lambda_i \cdot b_i
\end{equation}

This two-stage architecture---additive scoring for exploration, multiplicative gating for validation---provides an optimal balance \cite{meister2020best,cohen2019empirical,lemons2022beam}.

\subsection{Agent-Guided Multimodal Query Decomposition and Fusion}

\subsubsection{Query Expand and Decomposition}

We add a Query Expansion (QE) module powered by GPT-4o, following recent state-of-the-art work on Generalized Query Expansion (GQE) \cite{gqe2024} and Multi-Query Video Retrieval (MQVR) \cite{mqvr2022}. These studies show that using large language models (LLMs) to create multiple query variations helps to reduce noise from imperfect captions and better match human judgment \cite{gqe2024review}.

\begin{figure}[t]
\centering
\includegraphics[width=0.7\columnwidth]{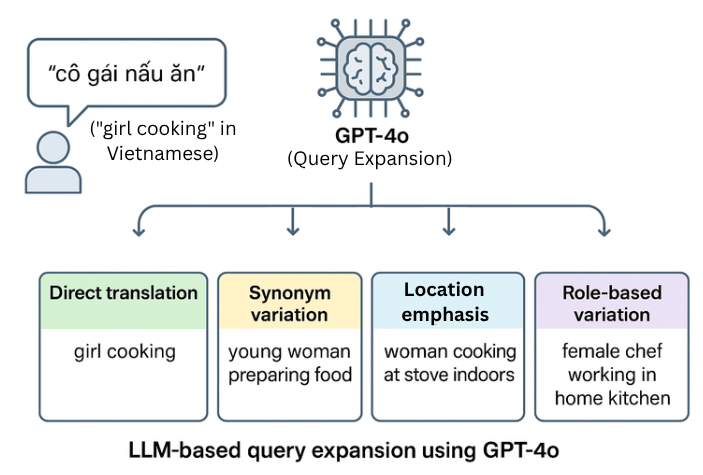}
\caption{LLM-based query expansion using GPT-4o.}
\label{fig:gptexpand}
\end{figure}

In our system, the QE module takes an original query and generates $N$ new variations (default $N=4$). These new queries focus on visual descriptions in English. The expansion follows two strict rules ( see Figure 3 ): Direct translation is required: The first expanded query is always a direct English translation of the original query to preserve meaning, as English ensures better embedding performance from our models. Original meaning is preserved: The variations may alter the visual angle, setting, or description style, but must not introduce new objects or actions not present in the original query. After expansion, all queries are embedded using SigLIP and BEiT-3, then reranked and searched in parallel.

\subsubsection{Modality Routing}

To address multimodal retrieval effectively, we leverage the reasoning capability of modern LLMs to construct an intelligent query coordinator. Instead of exhaustively querying all modalities or relying on a separately trained routing model, the agent infers the user's intent, decomposes the query when necessary, and dynamically assigns weights to the most relevant modalities. This enables efficient retrieval while avoiding the limitations of traditional approaches:

\textit{No exhaustive search:} Unlike late-fusion methods \cite{atrey2010multimodal}, our system does not blindly query all modalities, reducing computational overhead. \textit{No routing model training:} It eliminates the need for labeled routing data or specialized training pipelines \cite{wang2023laff}. \textit{No complex joint fusion:} It avoids costly multimodal embedding alignment required by joint fusion architectures \cite{lu2019vilbert}.

The routing is fully prompt-driven, with modality-specific heuristics embedded directly into prompt. Specifically, KIS (Visual Concepts) is used for elements that are \textit{seen} (e.g., actions, scenes, objects, colors), OCR (On-Screen Text) targets information that is \textit{read} from the screen (e.g., banners, numbers, jersey names), and ASR (Spoken Keywords) handles content that is \textit{heard}, such as speech-related keywords or lyrics (excluding generic action verbs like sing or speak).

When inference, the agent is guided to (1) detect modality-specific cues, (2) assess their distinctiveness in identifying relevant frames, (3) assign appropriate weights, and (4) explain its choice with a short reason for clarity. For example, given the query ``Cristiano Ronaldo scoring a goal'', the agent emphasizes KIS for the visual action of scoring, assigns moderate weight to OCR for the player's jersey name, and gives low weight to ASR, as commentary is generic.

\section{Experiments}
\subsection{Experimental Setups}
We evaluated our system on the AI Challenge 2025 dataset, which contains almost 1,500 videos and more than 200 GB of multimodal data. The evaluation followed the official competition setting and focused on three main tasks: Knowledge-based Image Search (KIS)~\cite{VBSChallenge} and Visual Question Answering (VQA)~\cite{VBSChallenge}, and Temporal Retrieval and Knowledge Extraction (TRAKE) is the task of locating temporally relevant video segments and extracting the associated factual or contextual knowledge needed to answer a query. Unlike standard retrieval tasks that focus on whole videos or static images, TRAKE requires the system to (1) identify the precise time interval in which the evidence appears, (2) interpret the temporal relations between events, and (3) extract or infer key information grounded in the video timeline to produce an accurate, knowledge-based response.. Our system is built on a layered retrieval-then-rerank architecture with an agent-based query decomposition module, as described in Section~3. For all experiments, we used a fixed set of key hyperparameters: Query Expansion expands each user query into \textbf{4} related variants; the first-stage retriever (BEiT-3 + SigLIP) selects the top 100 candidates which are then reranked by BLIP-2 for more accurate matching; and the temporal decay coefficient is set to 0.01 with a beam width of 8 for temporal search.

\subsection{Quantitative Results}

 Our method achieved a final score of \textbf{76.4/88}, ranking among the top-performing teams and advancing to the final round. This demonstrates that our layered retrieval--rerank architecture and agent-based decomposition strategy are effective in a challenging large-scale multimodal search setting.

Table~\ref{tab:basic-table}: summarizes the detailed scores across the three qualification rounds. The system showed consistent performance improvements over time, reaching the maximum score in Round~3.

\begin{table}[h]
\centering
\caption{Qualification scores across all rounds.}
\label{tab:basic-table}
\begin{tabular}{lccc}
\hline
\textbf{Round} & \textbf{Score} & \textbf{Max} & \textbf{Percentage} \\
\hline
Round 1 & 19.8 & 23 & 86.1\% \\
Round 2 & 26.6 & 30 & 88.6\% \\
Round 3 & 30   & 35 & 85.7\% \\
\hline
\textbf{Total} & \textbf{76.4} & \textbf{88} & \textbf{86.8\%} \\
\hline
\end{tabular}
\end{table}

To better understand the evaluation structure, Table~\ref{tab:query_distribution} reports the distribution of queries by task type and round.  
Round~1 and Round~2 contain a higher proportion of KIS queries, while VQA and TRAKE queries appear more sparsely but require deeper reasoning.  
The final round (Round~3) also contains the most KIS queries, reflecting its difficulty and the need for robust retrieval capabilities.

\begin{table}[h]
\centering
\caption{Distribution of evaluation queries across tasks and rounds.}
\label{tab:query_distribution}
\begin{tabular}{lccc}
\hline
\textbf{Round} & \textbf{KIS} & \textbf{VQA} & \textbf{TRAKE} \\ 
\hline
1 & 17 & 3 & 3 \\
2 & 26 & 2 & 2 \\
3 & 29 & 4 & 2 \\
\hline
\textbf{Total} & \textbf{72} & \textbf{9} & \textbf{7} \\
\hline
\end{tabular}
\end{table}
\begin{figure}
\centering
\includegraphics[width=0.95\columnwidth]{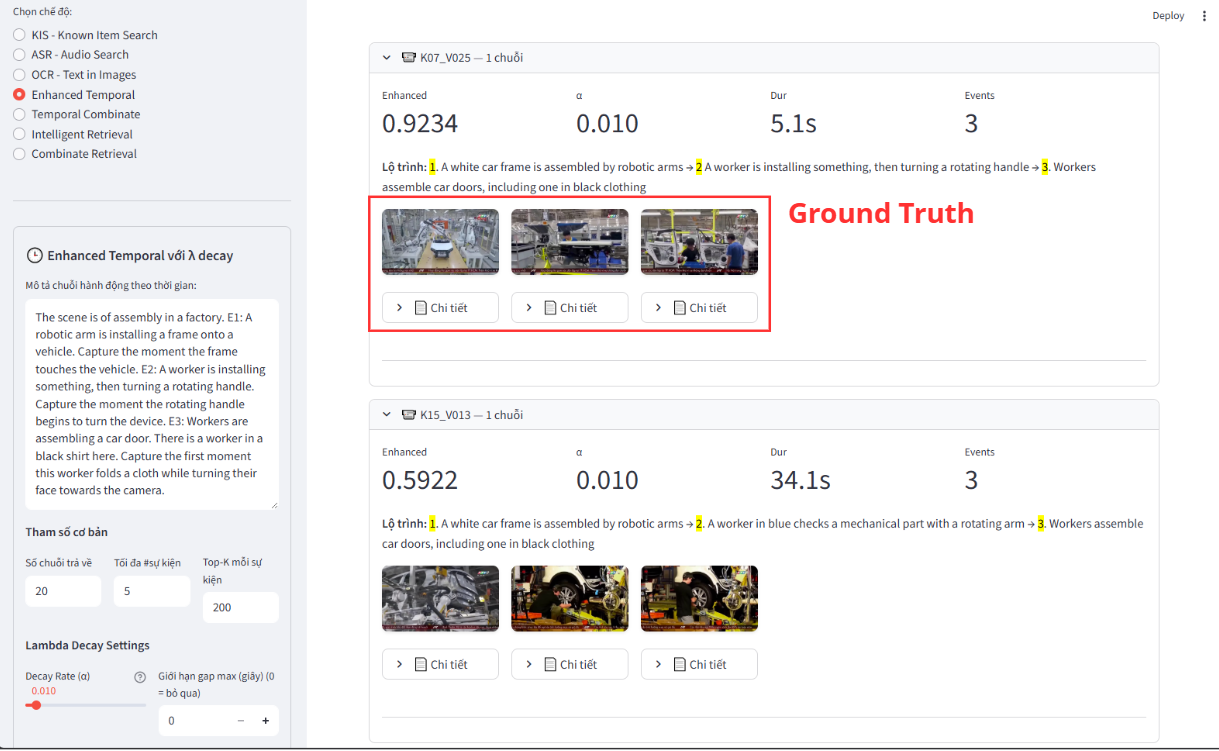}
\caption{Temporal coherence effectiveness.}
\end{figure}
\begin{figure*}[t]
\centering
\includegraphics[width=1.7\columnwidth]{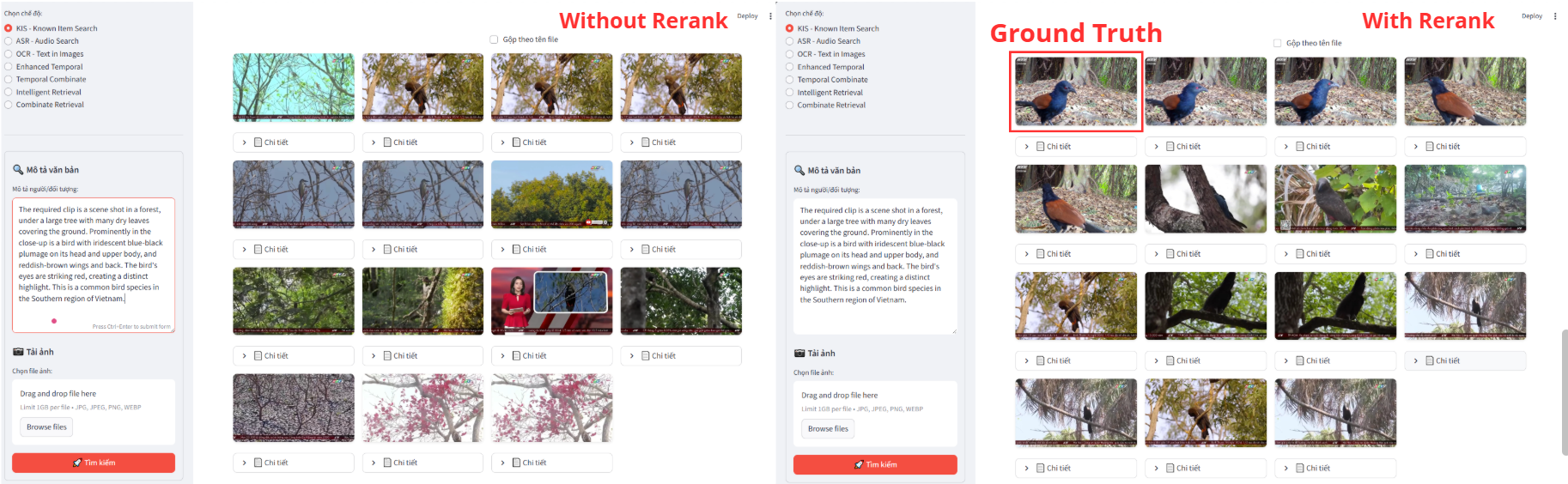}
\caption{\textbf{Example of our dual-stage retrieval pipeline.} The first-stage retriever (left) returns diverse but mostly irrelevant candidates and misses the Ground Truth. With the second-stage BLIP-2 reranker (right), the correct frame is surfaced at the top, showing the importance of fine-grained reranking for accurate temporal retrieval.}
\end{figure*}
\begin{figure*}[t]
\centering
\includegraphics[width=1.7\columnwidth]{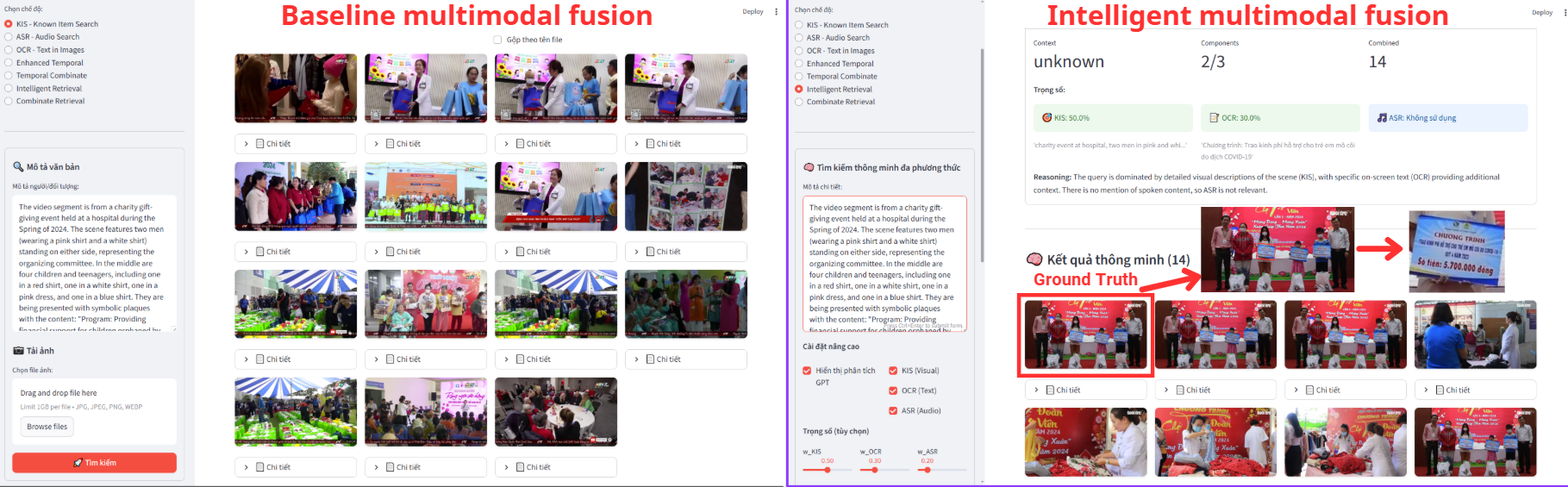}
\caption{\textbf{Comparison between baseline and intelligent multimodal fusion systems.} The baseline model retrieves visually similar but incorrect scenes due to the lack of contextual understanding. In contrast, the intelligent, Agent-guided system dynamically weights modalities—favoring OCR cues over visuals—to accurately locate the true target segment.}
\end{figure*}

\subsection{Qualitative Results}

\subsubsection{Cascaded Dual-Embedding Effectiveness}

This analysis shows a critical distinction: the baseline broad-search system (``Without Rerank'') fails. As shown on the left, while this initial stage retrieves a wide variety of candidates (high coverage), it cannot find the correct result. The Ground Truth (the blue bird) is absent from the top results, which are instead filled with irrelevant or visually dissimilar frames, such as a person in a red shirt and pink flowers. This demonstrates a clear lack of precision.

In contrast, our full system using a second-stage reranker (``With Rerank'') succeeds precisely where the baseline fails. By applying a more sophisticated model (such as BLIP-2) to do fine-grained image-text matching on the initial candidates, the system intelligently re-evaluates and re-scores the retrieved set. It is this strategic reranking that allows the system to correctly identify the true semantic match, pushing the Ground Truth frame to the top-ranked position.

This clearly demonstrates that the cascaded dual-stage pipeline (a broad search for coverage followed by a powerful reranker for precision) is the key factor in achieving optimal retrieval accuracy and overcoming the severe limitations of a single-stage search ( see Figure 4 ).

\subsubsection{Multimodal Fusion Effectiveness}

This analysis reveals a critical distinction: the baseline visual-only system fails. It retrieves many look-alike scenes (e.g., ``red background, children, holding a sign''), it cannot pinpoint the correct result due to a lack of contextual information. Its inability to perform multimodal fusion (integrating ASR and OCR) leaves it unable to distinguish between dozens of nearly identical visual candidates.

In contrast, our full system succeeds precisely where the baseline fails. By employing an Agent to decompose the query and guide the fusion process, the system intelligently assigns a higher weight to the OCR modality ($w_{\text{ocr}} \approx 0.7$) than to the visual features ($w_{\text{vis}} \approx 0.4$). Prioritizing the key text on the sign (``Program: Financial Support...'') lets it find and rank the correct scene.

This clearly demonstrates that the intelligent, Agent-guided integration of modality-specific information (like OCR) is the key factor in achieving optimal retrieval accuracy and overcoming the severe limitations of a visual-only approach ( see Figure 5 ).

\subsubsection{Temporal Coherence Effectiveness}

The system correctly identifies the actions in order: ``A white car frame is assembled by robotic arms'' $\rightarrow$ ``A worker is installing something, then turning a rotating handle'' $\rightarrow$ ``Workers assemble car doors, including one in black clothing'', with a short duration of 5.1 seconds and achieving a high score of 0.9234. The timeline visually confirms this sequence through three representative keyframes.

By applying Enhanced Temporal search with $\lambda$ decay (using $\lambda=0.010$ as specified in the settings), the system is able to prioritize this compact sequence. It successfully penalizes fragmented or lengthy results, such as the second candidate which had a much longer duration of 34.1s and a lower score. As a result, the system retrieves the meaningful narrative sequence of the factory assembly rather than returning isolated, out-of-order frames ( see Figure 6 ).

\section{Future Work}

First, we will improve how the system connects temporal and semantic information. Instead of relying solely on exponential decay to preserve temporal proximity, we plan to incorporate models capable of understanding semantic relationships between events. This enhancement will enable the system to produce clearer and more meaningful storylines rather than returning isolated frames that merely occur close in time.

Next, we aim to integrate a multimodal language model capable of processing both images and audio to generate unified captions for video segments. These captions will then be converted into embeddings for retrieval. This approach will allow the system to construct a more coherent and semantically rich index, instead of treating each modality independently.

Finally, we will upgrade the user interface to make it more intuitive and user-friendly. In parallel, user feedback will be incorporated into the system so it can dynamically adjust its fusion strategies over time, gradually improving the accuracy and relevance of its retrieval results.

\section{Conclusion}

We present a unified multimodal video retrieval system with three key contributions: (1) a cascaded dual-embedding architecture balancing scalability and precision through dual encoders and reranking, (2) temporal reasoning with exponential decay ($\lambda_i = e^{-\alpha \cdot \Delta t_i}$) constructing coherent event sequences via beam search, and (3) Agent-guided query processing enabling automatic decomposition and adaptive multimodal fusion.

Qualitative analysis demonstrates effective handling of ambiguous queries, temporally coherent retrieval, and dynamic modality adaptation. Future directions include user feedback integration, hierarchical temporal modeling, and large-scale benchmark evaluation on TRECVID and VBS. This work advances practical interactive video search for modern multimodal content ecosystems.

\bibliography{aaai2026}

\end{document}